\documentclass[10pt,twocolumn,letterpaper]{article}

\usepackage{iccv}
\usepackage{times}
\usepackage{epsfig}
\usepackage{graphicx}
\usepackage{amsmath}
\usepackage{amssymb}
\usepackage{multirow}
\usepackage{algorithmic}
\usepackage{algorithm}
\usepackage{color}
\usepackage{colortbl}
\newcommand{\tabincell}[2]{\begin{tabular}{@{}#1@{}}#2\end{tabular}}
\usepackage{url}
\usepackage{xspace}
\usepackage{xcolor}
\usepackage{blindtext}
\usepackage{indentfirst}
\usepackage{ragged2e}
\usepackage{makecell}
\usepackage{diagbox}
\usepackage{bm}
\usepackage{booktabs} 
\usepackage[misc]{ifsym}
\definecolor{mygray}{RGB}{208, 206, 206}
\definecolor{myorange}{RGB}{237, 125, 49}
\definecolor{myblue2}{RGB}{157, 193, 230}
\definecolor{myblue3}{RGB}{76, 114, 181}

\newcommand{\blue}[1]{\textcolor{blue}{#1}}

\usepackage{caption}
\usepackage{graphicx}
\usepackage{booktabs}
\usepackage{blindtext}

\usepackage[pagebackref=true,breaklinks=true,letterpaper=true,colorlinks,bookmarks=false]{hyperref}

\iccvfinalcopy 

\def\iccvPaperID{1553} 

\begin{document}

\title{
%
Ranking Models in Unlabeled New Environments
}

\author{Xiaoxiao Sun, Yunzhong Hou, Weijian Deng, Hongdong Li, Liang Zheng\\
Australian National University\\
{\tt\small \{first name.last name\}@anu.edu.au}
}

\maketitle
\thispagestyle{empty}

\begin{abstract}

Consider a scenario where we are supplied with a number of ready-to-use models trained on a certain source domain and hope to directly apply the most appropriate ones to different target domains based on the models' relative performance. 
Ideally we should annotate a validation set for model performance assessment on each new target environment, but such annotations are often very expensive. Under this circumstance, we introduce the problem of ranking models in unlabeled new environments. 
For this problem, we propose to adopt a proxy dataset that 1) is fully labeled and 2) well reflects the true model rankings in a given target environment, and use the performance rankings on the proxy sets as surrogates.  
We first select labeled datasets as the proxy.
Specifically, datasets that are more similar to the unlabeled target domain are found to better preserve the relative performance rankings.
Motivated by this, we further propose to search the proxy set by sampling images from various datasets that have similar distributions as the target. 
We analyze the problem and its solutions on the person re-identification (re-ID) task, for which sufficient datasets are publicly available, and show that a carefully constructed proxy set effectively captures relative performance ranking in new environments. Code is available at \url{https://github.com/sxzrt/Proxy-Set}.
\end{abstract}

\section{Introduction}
In real-world applications, it is not uncommon to see models trained on the source domain (hereafter called source models) directly applied to unlabeled new target environments (hereafter called target domains) at the price of employing some unsupervised domain adaptation (UDA) techniques \cite{ganin2016domain,long2015learning,hoffman2018cycada}. Assume that one has access to a pool of source models and can choose appropriate ones. Under this context, it is desirable to obtain the relative performance of different models on the target domain without having to annotate data in the target environment. 

\begin{figure}[t]
\begin{center}
	\includegraphics[width=0.95\linewidth]{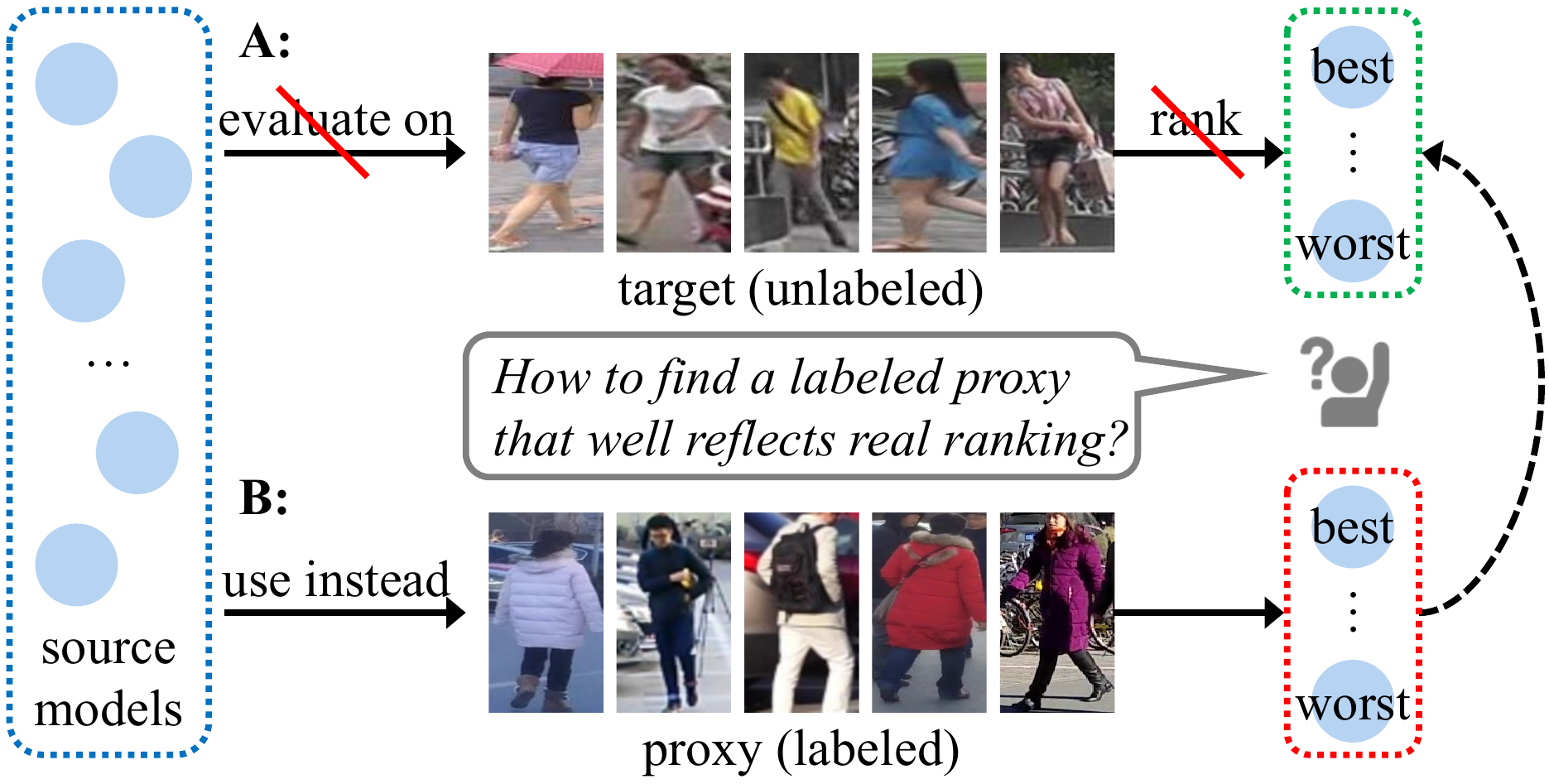}
\end{center}
\vspace{-1mm}
\caption{ Illustration of the proposed problem and a general solution. Given various models (\textcolor{myblue3}{blue} circles) trained on source data (hereon denoted as source models) and an unlabeled target domain, we aim to rank them and find the best one for direct deployment to the target. \textbf{A}: Without access to image labels, this objective is unlikely to be achieved using only the target data. \textbf{B}: We find a proxy to rank the model performance and use this (\textcolor{red}{red}) ranking as a surrogate. 
Specifically, this proxy should 1) be fully labeled and 2) well reflect the (\textcolor{green}{green}) true ranking on the target. 
}
\vspace{-1mm}
\label{fig:fig1}
\end{figure}

To find the appropriate models, we usually evaluate each individual model on a labeled partition (\eg, a validation set) of the target environment and rank them to find the best one (see Fig. \ref{fig:fig1} \textbf{A}). However, annotation is often expensive to obtain, and it becomes prohibitive if we consider data labeling for every new application scenario. 
As such, an interesting question arises: can we estimate model rankings in new environments in the absence of ground truth labels? 

In this work, we aim to find a proxy (or surrogate) to rank the models in answer to the aforementioned question.  Specifically, we focus on the person re-identification (re-ID) task, which aims to retrieve persons of the same identity across multiple cameras. 
For this problem, it is desirable that the proxy can provide similar model rankings, since a target validation set is difficult to acquire in practice.  To this end, the proxy should satisfy: 1) have labels for evaluation and 2) well reflect the true model rankings (see Fig.~\ref{fig:fig1} \textbf{B}).

For the first requirement (labels), we can either use the target dataset with \textit{pseudo} labels, or other labeled datasets. However, due to the nature of pseudo labels, some of them might not be accurate. Existing works find that the inaccurate pseudo labels greatly influence the model accuracies when used in training \cite{fan2018unsupervised}. We suspect such inaccurate pseudo labels may even do more harm when used for evaluation. As such, we consider using labels that are \emph{real} and not from the target domain. 

For the second requirement (a good reflection of the true ranking), we should consider the target data distribution. If we intuitively use the model rankings on the source domain (assuming a labeled source validation set) for the ranking estimation, we might find them to be very different from the target rankings. This can often be attributed to the distribution difference. For example, one model may outperform another in a certain scenario, but their performances could be dis-similar or even reversed in a different scenario. Therefore, in order to obtain accurate model rankings on the target domain, target data distribution should be considered.

We explore proxy sets that meet these two requirements. \textbf{First}, we use existing datasets, where the labels of IDs are available. It could be the source, an arbitrary dataset other than the source or target, or a composite one. This allows us to conveniently compute model accuracies using its labels. \textbf{Second}, the proxy is close to the target distribution in terms of two distribution difference measurements: Fr\'{e}chet Inception Distance (FID)~\cite{heusel2017gans} and feature variance gap \cite{dubey2018maximum,jonsson1982some}. 
This is based on our observation that datasets more similar to the target domain (\ie, small FID and small variance gap) are more likely to form better proxies. This observation shares a similar spirit with some key findings in domain adaptation that reduced domain gap can benefit model training. Yet we derive it from a different viewpoint, \ie, the quality of a proxy set for performance ranking. 

These two measurements are further investigated in a dataset search procedure. An image pool is collected from existing datasets and is partitioned into clusters. Images are sampled from each cluster with a probability proportional to the similarity (FID and variance gap) between the cluster and the target, forming the proxy. Overall, this paper contains the following main points.

\begin{itemize}
  \item We study a new problem: ranking source model performance on an unlabeled target domain. 
  \item We propose to use a labeled proxy that can give us a good estimation of model ranking. It is constructed via a search process such that the proxy data distribution is close to the target.
  \item Experiment verifies the efficacy of our method, and importantly, offers us insights into dataset similarities and model evaluation.
\end{itemize}

\section{Related Work}
\label{sec:related work}

\textbf{Unsupervised domain adaptation (UDA)} is a commonly used strategy to improve source model performance on the target domain where no labeling process is required. This objective can be implemented on the feature level \cite{long2015learning}, pixel level \cite{zhu2017unpaired, deng2018similarity}, or based on pseudo labels \cite{fan2018unsupervised,zhong2019invariance,song2020learning}. 
While the goal of UDA is to learn an effective model for target scene, we aim to compare the performance of different models that are directly transferred to the target domain. 


\textbf{Predicting model generalization ability.} Our work is also related to this area, where model generalization error on unseen images is estimated. 
Some work predicts the generalization gap using the training set and model parameters \cite{arora2018stronger,corneanu2020computing,jiang2018predicting,neyshabur2017exploring}. For example, Corneanu \textit{et al}.~\cite{corneanu2020computing} use the persistent topology measures to predict the performance gap between training and testing errors.
There are also works aiming to predict accuracy on unlabeled test samples based on the agreement score among predictions of several classifiers \cite{madani2004co,platanios2016estimating,platanios2017estimating,donmez2010unsupervised}. Platanios \etal \cite{platanios2017estimating} use a probabilistic soft logic model to predict classifier errors. 
Recently, Deng \etal~\cite{deng2020labels, Deng:ICML2021} attempt to estimate classifier accuracy on various unlabeled test sets. Our work differs from the above works. We study a new problem: ranking different models in an unlabeled test domain.

\textbf{Learning to simulate synthetic data.} The objective of this area is to bridge the gap between the synthetic and real-world images by optimizing a set of parameters of a surrogate function that interfaces with a synthesizer~\cite{yao2019simulating, xue2020learning, kar2019meta}. It can be used to make customized data but needs to utilize specific engines and 3D models similar to the target object, which is not often accessible. Some recent works~\cite{li2017webvision,yan2020neural} search a dataset from websites or data server for model training. Inspired by them, we attempt to search a proxy set with annotated data to rank models for the target domain. 

\begin{figure*}[t]
\begin{center}
	\includegraphics[width=\linewidth]{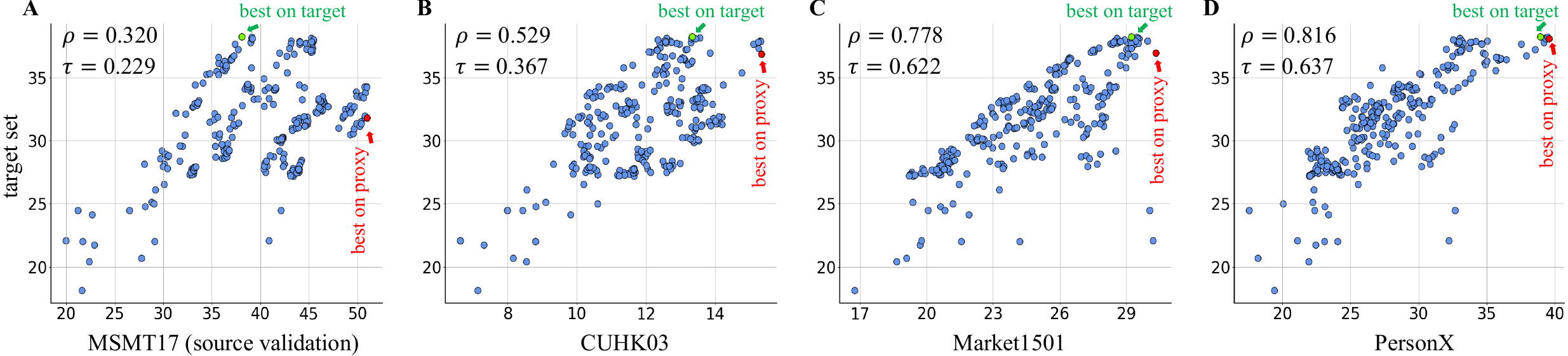}
\end{center}
\caption{Correlation of model accuracies (mAP, \%) on a given target set \vs proxy sets. Specifically, we train models with MSMT17 as the source domain and DukeMTMC-reID as the target. Four proxy choices are studied, \ie, from left to right: \textbf{A.} source (MSMT17) validation set, \textbf{B.} CUHK03, \textbf{C.} Market-1501, and \textbf{D.} PersonX. For each model (\textcolor{myblue3}{blue} circles), we evaluate its mAP scores on the target test set and the proxy set, which are then used to plot a 2-D correlation sub-graph. 
For each sub-graph, we use Spearman’s Rank Correlation ($\rho$)~\cite{spearman1961proof} and Kendall's Rank Correlation ($\tau$)~\cite{kendall1938new} to measure the correlation between the two sets of mAP values. A higher absolute value of $\rho$ (or $\tau$) indicates a stronger correlation. Also shown are the best models on the target (\textcolor{green}{green} circle) and the proxy (\textcolor{red}{red} circle). We clearly see that source is a relatively poor proxy ($\rho=0.320$, $\tau=0.229$), while PersonX ($\rho= 0.816$, $\tau=0.637$) and Market-1501 ($\rho=0.778$, $\tau=0.622$) are much better choices. Aside from the increased correlation (from left to right), we also find that the best models on the target and the proxy are getting closer. It indicates that the best model on the target is more likely to be the best one on the proxy (with a smaller error). 
All the above correlation coefficients have very high statistical significance due to their p-value $ < 0.001$. 
}
\label{fig:fig2}
\end{figure*}

\textbf{Learning to rank} has been studied in the fields of information retrieval~\cite{qin2010letor,wan2015online,hu2008multiple}, data mining~\cite{joachims2017unbiased, cakir2019deep} and natural language processing~\cite{tay2017learning, goldberg2017neural}. In general, given a query, the goal is to learn to rank data from the collection and return the top-ranked data. In computer vision, learning to rank is studied in content-based image retrieval~\cite{feng2015learning,hu2008multiple} and metric learning~\cite{ge2018deep,cakir2019deep,liu2011learning}. These works are concerned with learning metrics so that related samples are mapped to be closer to the query than unrelated ones. While they work on the datum (image) level, our paper deals with model ranking, which is on the model level. 

\section{Problem and Baseline}
\label{sec:method}

\subsection{Problem Definition}
Let $\left \{ \mathbf{m}_{i} \right \}_{i=1}^{M}$ denote a set of $M$ models trained on source domain (we call them source models). $\mathbf{T}$ is a set of unlabeled images collected from the target domain for performance ranking. In order to find the best model for direct application on the target domain, ideally, we should rank the model performances on $\mathbf{T}$ after labeling all images. However, given the high annotation costs, this becomes a less appealing choice.
In this paper, we investigate whether it is possible to estimate the model accuracy ranking on the target domain (hereon denoted as the \textit{ground truth} accuracy ranking) without labeling the images in $\mathbf{T}$. 

Specifically, given an unlabeled target dataset $\mathbf{T}$ and models $\left \{ \mathbf{m}_{i} \right \}_{i=1}^{M}$, we aim to find a labeled proxy set $\mathbf{P}$ whose performance ranking well represents the ground truth accuracy rankings on $\mathbf{T}$. We therefore formulate the goal of this problem as, find  $\mathbf{P}$,
\begin{equation}
\begin{aligned}
\text{s.t. } \quad & rank\left(\left \{ \mathbf{m}_{i} \right \}_{i=1}^{M}, \mathbf{P}\right) \rightarrow  rank\left(\left \{ \mathbf{m}_{i} \right \}_{i=1}^{M}, \mathbf{T}\right),
\end{aligned}
\end{equation}
where $rank\left(\cdot, \cdot\right)$ denotes the performance ranking of certain models on a certain dataset.

For each proxy dataset, we use the model accuracies to create a performance ranking, and evaluate the quality of the proxy set as its ranking correlation with the ground truth accuracy ranking on target domain. To quantitatively evaluate the quality of a proxy, we use two rank correlation coefficients: Spearman’s Rank Correlation $\rho$~\cite{spearman1961proof}, and Kendall’s Rank Correlation $\tau$~\cite{kendall1938new}. 
Both $\rho$ and $\tau$ fall into the range of $\left[-1,1\right]$, and a higher absolute value indicates a stronger correlation between rankings, \ie, $rank\left(\left \{ \mathbf{m}_{i} \right \}_{i=1}^{M}, \mathbf{P}\right)$ and $rank\left(\left \{ \mathbf{m}_{i} \right \}_{i=1}^{M}, \mathbf{T}\right)$ in our problem. Accordingly, a lower absolute value of the correlation scores (with $0$ being the lowest) indicates weak (or no) correlation.

\subsection{Baseline: Individual Datasets as Proxy}
\label{sec:intutive solution}

\textbf{Source validation set as proxy.}~We first study the relationship between model performance on the source (MSMT17~\cite{wei2018person}) validation set (we use the test partition in the absence of validation) and target (DukeMTMC-reID~\cite{zheng2017unlabeled,ristani2016MTMC}) test set.~Specifically,~280 re-ID models trained on MSMT17 are considered, which are shown by blue circles in Fig.~\ref{fig:fig2}.~We plot these circles according to their accuracies on the proxy (MSMT17) and the target (DukeMTMC-reID). We only report mean average precision (mAP) here and omitted rank-1 precision since both metrics share a very similar trend. The rank correlation coefficients are: $\rho=0.320$ and $\tau=0.229$, indicating a weak rank correlation~\cite{akoglu2018user,marshall2016statistics} between proxy and target. As an intuitive understanding, the best model according to the proxy (source validation) has mAP $5.5\%$ lower than the best one on the target set. We also witness similar phenomena using different source and target datasets. These results show that the source is a less appealing choice for proxy. 


\textbf{Other datasets as proxy.}~An annotated dataset from another domain can be a proxy, too. For example, when using MSMT17 and DukeMTMC-reID as source and target, respectively, a third dataset, Market-1501 \cite{zheng2015scalable} can serve as a target proxy. There are also other options readily available, such as PersonX~\cite{sun2019dissecting} and RandPerson~\cite{wang2020surpassing} (see Fig.~\ref{fig:fig2} \textbf{B-D}). 
When compared with source validation set (MSMT17, Fig.~\ref{fig:fig2} \textbf{A}), these datasets consistently achieve higher ranking correlations. 
For example, when using CUHK03, Market-1501, and PersonX as the proxy, we obtain Spearman's $\rho$ of $0.529$, $0.778$, and $0.816$, and Kendall's $\tau$ of $0.367$, $0.622$, and $0.637$, respectively. These numbers are consistently higher than those calculated from using the source as proxy. Meanwhile, the correlation coefficients suggest that the Market-1501 and PersonX are ``moderate to strong'' rank correlated with the target test set on model performance~\cite{akoglu2018user,marshall2016statistics}.
When using different source and target combinations, we also find that these datasets from different domains form better proxies when compared to corresponding source validations. 
In this case, unless specified, we do not use the source validation as proxy in our further experiments. See Section~\ref{sec:discussion of problem} for more discussions.

\section{Method: Search a Proxy Set}

\subsection{Motivation} \label{sec:discussion_motivation}

When using different datasets (other than source and target) as proxy, we find some proxy sets to have higher quality (higher correlations with ground truth ranking) than others. Interested in what causes such proxy quality differences, we further investigate the potential reasons. Inspired by works in domain adaptation~\cite{deng2018similarity,zhu2017unpaired}, we examine the distribution difference between proxy set $\mathbf{P}$ and the target set $\mathbf{T}$. 
Specifically, we measure the distribution difference via two metrics, Fr\'{e}chet Inception Distance (FID) \cite{heusel2017gans} and feature variance gap \cite{jonsson1982some}. $\mathrm{FID}(\mathbf{T},\mathbf{P})$ measures the domain gap between the proxy set $\mathbf{P}$ and target set $\mathbf{T}$. On the other hand, feature variance gap measures how similar two data distributions are in terms of diversity and variation. We compute feature variance gap as the absolute difference between feature variance of $\mathbf{P}$ and $\mathbf{T}$,
\begin{equation}
\mathrm{V_{gap}}(\mathbf{P},\mathbf{T})=\left | v(\mathbf{P}) - v(\mathbf{T}) \right |,
\end{equation}
where $v\left(\cdot\right)$ computes the variance.
Notably, to calculate FID and $\mathrm{V_{gap}}$, we use Inception-V3~\cite{szegedy2016rethinking} pre-trained on ImageNet to extract image features.

\begin{figure}
\begin{center}
	\includegraphics[width=\linewidth]{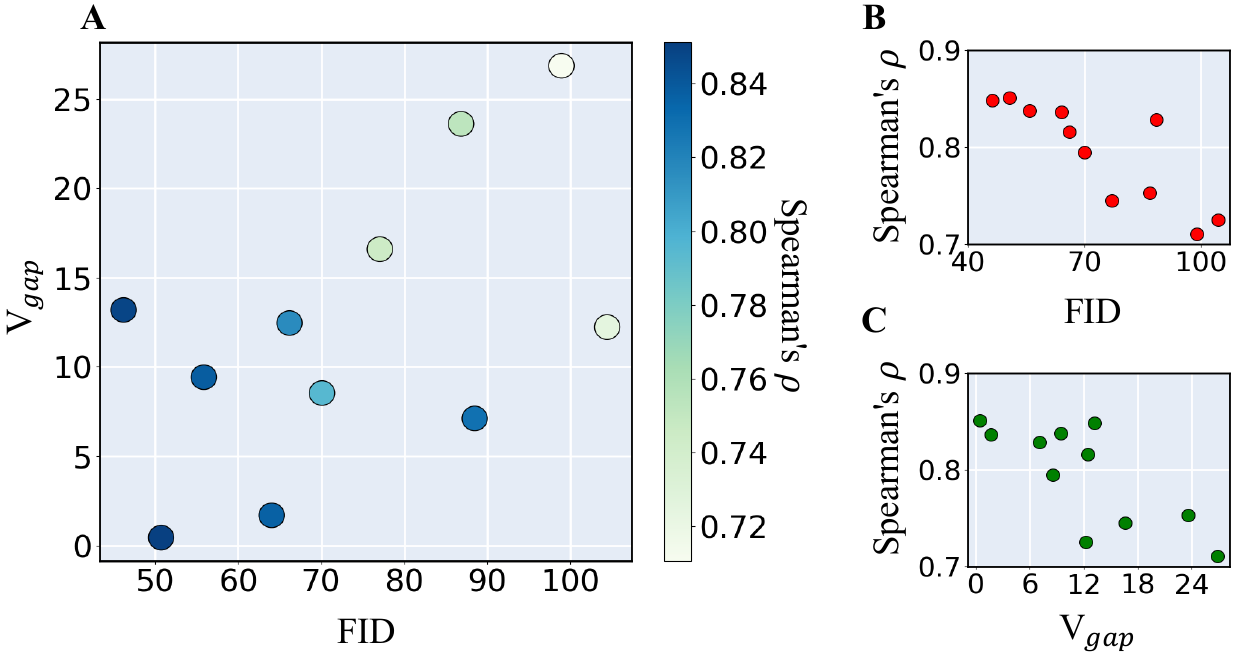}
\end{center}
\caption{Relationships between FID, variance gap and the proxy set quality (evaluated using $\rho$).
\textbf{A}: Influence of FID and variance gap on the proxy set quality. 
\textbf{B}: FID \vs proxy set quality. There is a very strong negative correlation ($-0.88$) with a very high statistical significance ($\text{p-value}<0.001$) between them. \textbf{C}: Variance gap \vs proxy set quality. They have a relatively strong negative correlation ($-0.65$) with a high statistical significance ($\text{p-value}<0.05$). 
These three sub-figures verify that both FID and variance gap affect the proxy quality.
}
\label{fig:fid}
\end{figure}

Using these two metrics, we further show the relationships between FID, $\mathrm{V_{gap}}$, and proxy quality (correlations with the ground truth ranking). As shown in Fig.~\ref{fig:fid} \textbf{A}, we can spot an overall trend that smaller FID and $\mathrm{V_{gap}}$ values often accompany higher proxy quality (ranking correlation coefficients). 
Moreover, as from Fig.~\ref{fig:fid} \textbf{B} and \textbf{C}, there also exist relatively strong correlations between either of the two metrics and the quality of proxy sets. 

These experiments show that there might exist a proxy set of even better quality if it is composed of images whose distributions are more similar to the target (in terms of FID and $\mathrm{V_{gap}}$). Motivated by this observation, we explore how to create a proxy set by searching images in next section.

\subsection{The Search Algorithm}\label{sec:search}
\label{sec:search a set}
\begin{figure}
\begin{center}
	\includegraphics[width=\linewidth]{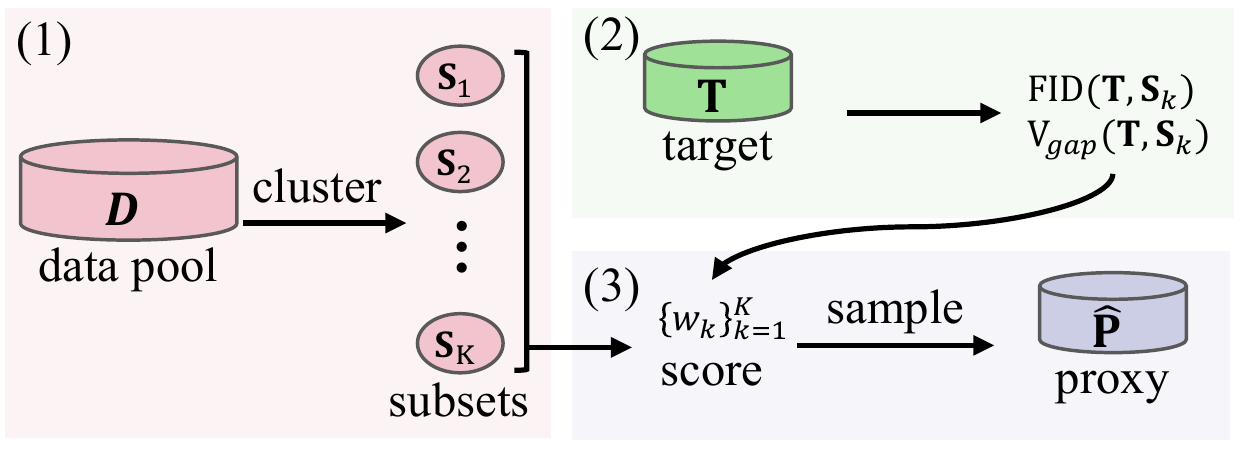}
\end{center}
\caption{Three steps in our proxy searching process. First, we cluster the data pool into $K$ clusters; second, we compute the distribution differences between clusters and the target; third, we calculate the sampling scores and compose a proxy set accordingly. 
}
\label{fig:pip}
\end{figure}

Given a data pool $\mathbf{D}$ that includes multiple datasets (other than the source and the target) and an unlabeled target set $\mathbf{T}$, it is our goal to sample data from $\mathbf{D}$ and compose a proxy set $\hat{\mathbf{P}}$ that has small $\mathrm{FID}(\mathbf{T},\hat{\mathbf{P}})$ and $\mathrm{V_{gap}}(\mathbf{T},\hat{\mathbf{P}})$. Based on the findings in Section~\ref{sec:discussion_motivation}, we believe this can lead to a high quality proxy set for target domain. 
As shown in Fig.~\ref{fig:pip}, we go through the following three steps in our proxy searching approach: 

First, we cluster the data pool $\mathbf{D}$ into $K$ subsets $\left \{ \mathbf{S}_k \right \}_{k=1}^{K}$. To this end, we average all image features that belong to the same identity, and use this ID-averaged feature to represent all corresponding images. We then use $k$-means~\cite{likas2003global,malinen2014balanced} to cluster the ID-averaged features into $K$ groups, and construct $K$ subsets by including all images of the corresponding IDs that are in that group.  

Second, we calculate the $\mathrm{FID}(\mathbf{T}, \mathbf{S}_k)$ and $\mathrm{V_{gap}}(\mathbf{T}, \mathbf{S}_k)$ between each subset and the target set $\mathbf{T}$. 

Lastly, we calculate a sampling score $\left \{w_{k} \right \}_{k=1}^{K}$ for each subset, and then assign a probabilistic weighting for each ID and sample ID form the data pool $\mathbf{D}$ based on the weightings. Specifically, we calculate the sampling score based on $\left \{ \mathrm{FID}(\mathbf{T}, \mathbf{S}_k) \right \}_{k=1}^{K}$ and $\left \{ \mathrm{V_{gap}}(\mathbf{T}, \mathbf{S}_k) \right \}_{k=1}^{K}$. We take the negative of FID and variance gap values when calculating the sampling scores according to the negative correlations between their values and the proxy quality (see Fig.~\ref{fig:fid}). 
The sampling score is written as,
\begin{equation}\small
\label{eq:4}
\begin{aligned}
\left \{w_{k} \right \}_{k=1}^{K}&=\lambda softmax(\left \{-\mathrm{FID}(\mathbf{T}, \mathbf{S}_k) \right \}_{k=1}^{K}) \\ &+ (1-\lambda)softmax(\left \{-\mathrm{V_{gap}}(\mathbf{T}, \mathbf{S}_k) \right \}_{k=1}^{K}),
\end{aligned}
\end{equation}
where $softmax(\cdot)$ denotes the softmax function, and $\lambda\in [0,1]$ is a weighting factor. $\lambda=0$ or $1$ represents only using FID or variance gap to calculate sampling score. Based on the sampling scores of clusters, each ID of each cluster is assigned a probabilistic weighting $\frac{w_{k}}{\left | \mathbf{S}_k \right |}$. Here,  $\left | \mathbf{S}_k \right |$ is the number of IDs of the cluster $\mathbf{S}_k$. The proxy set is constructed by sampling $N$ examples from the data pool $\mathbf{D}$ at a rate according to probabilistic weightings of IDs.

In addition, if the camera annotation of the target set is available, 
we can further split the searching process into $N$ steps for $N$ cameras in the target, and then combine the final results as the proxy set $\hat{\mathbf{P}}$. Specifically, we repeat the aforementioned procedure $N$ times (each camera once) to get $N$ proxy sets. 
Notably, if one identity is sampled multiple times, we keep only one copy of the images of that identities in the final proxy set. 
We believe such a task-specific design would be helpful as it aligns with the multi-camera matching nature of re-ID problems~\cite{zheng2016person}. 

\subsection{Discussion}
\label{sec:discussion of problem}
\textbf{Why is the source often a poor proxy?} Two reasons would explain the trend in Fig.~\ref{fig:fig2} \textbf{A}. First, in our experiment, there is a non-negligible domain gap between the source (\eg, MSMT17) and target (\eg, DukeMTMC-reID).~A strong model capable of distinguishing between fine-grained classes on the source may lose such discriminative ability on the target due to their distribution difference. Second, the models may be more or less overfitting the source. It is shown in~\cite{kornblith2019better} that when pretrained on ImageNet~\cite{deng2009imagenet}, models that have higher accuracy on ImageNet yields superior accuracy on other classification tasks after fine-tuning. While there seem to be fewer overfitting issues with ImageNet pretrained models, the relatively small source datasets (\eg, MSMT17) in re-ID may cause overfitting, such that a good model on the source may be poor under a different environment.

\textbf{Distribution difference measurements.}
This paper computes sampling weights based on both FID and variance gap ($\mathrm{V_{gap}}$). Interestingly, the computation of FID also includes a diversity term between the two distributions, as it uses the covariance matrix. Nonetheless, in the experiment, we find only using either FID or variance gap leads to inferior results than them combined (see Fig.~\ref{fig:lambda}), which suggests both of them are indispensable. 
This suggests that the adopted feature variance gap could really benefit the searching process since it might provide a different angle for diversity difference measurement of data distribution. 

\textbf{Application scope.} %
The proposed problem and solution allow us, for example, to select the most suitable model for new environments. 
As shown in Fig.~\ref{fig:fig2} and later experiments, the selection process is fairly reliable.
For applications like object recognition, we require that the proxy have the same categories as the target and the source so that the source models can be evaluated. The number of such classification datasets is currently limited (see supplementary material).
For applications like person re-identification, we can leverage the abundant datasets available for proxy construction, because it is feasible to evaluate source models on proxy sets with completely different categories. 
In addition, since the proposed task is new and challenging, we currently focus on models that are directly applied to target data to avoid complicating the problem. As such, we do not consider UDA models~\cite{deng2018similarity,zhong2020learning} that include the target samples in training, but they are worth studying, and we will investigate these models in future works.

\begin{table}[t]\small
\begin{center}
    \setlength{\tabcolsep}{3.3mm}{
        \begin{tabular}{l|c|c|c} 
            \Xhline{1.2pt}
            \multicolumn{1}{l|}{Dataset}    &    \multicolumn{1}{c|}{\#ID} &     \multicolumn{1}{c|}{\#images} &    \multicolumn{1}{c}{\#ID in $\mathbf{D}$}\\
            \hline
      MSMT17  & 4,101 & 126,441  & 3,060    \\
       \hline
     DukeMTMC-reID & 1,812  & 36,411 & 702 \\
            \hline
      Market-1501  &  1,501 & 32,668 & 750  \\
            \hline
      CUHK03  &  1,467 & 13,164 & 700 \\
            \hline
      RAiD    &  43 & 1,264 & 43 \\
            \hline
       iLIDS &  119 & 476 & 119 \\
            \hline
       PKU-Reid   & 114 &1,824  & 114 \\
       \hline
 PersonX & 1,266 & 227,880 & 856 \\
  \hline
  RandPerson& 8,000 & 228,655 &  1,000 \\
  \hline
    UnrealPerson& 3,000& 120,000 &  800\\
            \Xhline{1.2pt}             
    \end{tabular}}
\end{center}
    \caption{Data pool composition. Seven real-world datasets and three synthetic datasets are considered. \#ID in $\mathbf{D}$ means the number of identities used in the data pool. } \label{table:Datasets}
\end{table}

\begin{table*}[t]\small
\begin{center}
    \setlength{\tabcolsep}{0.5mm}{
        \begin{tabular}{l| l | l |c c c c c c c |c c c c|c c} 
            \Xhline{1.2pt}
  \multicolumn{1}{c|}{\multirow{2}{*}{Source}} &   \multicolumn{2}{c|}{\multirow{2}{*}{Target}}  &\multicolumn{7}{c|}{Individual Dataset} & \multicolumn{4}{c|}{Other Method} & \multicolumn{2}{c}{Ours}\\ 
 \cline{4-16}
 \multicolumn{1}{c|}{} & \multicolumn{2}{c|}{}  & \scriptsize{CUHK03} & \scriptsize{Duke} & \scriptsize{Market} & \scriptsize{MSMT17} & \scriptsize{RandPerson} & \scriptsize{PersonX} & \scriptsize{UnrealPerson} & \scriptsize{Random} & \tiny{Attr. descent~\cite{yao2019simulating}} & \tiny{StarGAN~\cite{choi2018stargan}} & \tiny{pseudo-label~\cite{fan2018unsupervised}} & w/o cam & w/ cam \\ 
 \Xhline{1.2pt}
  \multirow{4}{*}{MSMT17} & \multirow{2}{*}{Duke}& 
  $\rho$ &  0.529 & -  & 0.778 &0.320& 0.775 & 0.816 & 0.837 & 0.725 & 0.756 & 0.700 & 0.789 & \textbf{0.858} & \blue{\textbf{0.882}}\\
  &  & $\tau$ & 0.367 & -&0.622 & 0.229& 0.602 & 0.637  & 0.655 & 0.537  &0.569 &0.518&  0.625& \textbf{0.713} &  \blue{\textbf{0.725}} \\
     \cline{2-16}
    &\multirow{2}{*}{Market}&   
      $\rho$ & 0.180 & 0.778& -& 0.335 & 0.803 & 0.874  & 0.854 &  0.643 &  0.638 &0.811 & 0.823 & \textbf{0.884} &  \blue{\textbf{0.912}}\\
     & & $\tau$ &  0.126 & 0.622 & - & 0.245 & 0.616 &  0.690 & 0.664 & 0.507 & 0.467 & 0.615&0.648 & \textbf{0.715} & \blue{\textbf{0.753}}\\
\Xhline{1pt} 
  \multirow{4}{*}{Market}  & \multirow{2}{*}{Duke}&
          $\rho$ & 0.374 & - & -0.119& 0.932& 0.905 &0.805& 0.933 & 0.713 &0.740&  0.848&0.899 &\textbf{0.939} &\blue{\textbf{0.950}} \\
      &  & $\tau$ & 0.260 & - & -0.048 & 0.790& 0.774& 0.626& 0.808 & 0.538 &0.551&  0.662&0.742 & \textbf{0.810} & \blue{\textbf{0.824}}\\
  \cline{2-16}
    & \multirow{2}{*}{MSMT17} &$\rho$ 
                  & 0.331 & 0.932 &-0.173 & -& 0.876 &  0.727 & 0.941 & 0.711 & 0.790& 0.807&0.846 & \textbf{0.949}& \blue{\textbf{0.958}}\\
      &  & $\tau$ & 0.254 & 0.790 & -0.092& -&0.705  & 0.548 & 0.817 & 0.553 & 0.612 &0.624 & 0.698& \textbf{0.822}& \blue{\textbf{0.829}}\\
\Xhline{1.2pt} 
    \end{tabular}}
\end{center}
\caption{Comparison of different proxy sets on different source-target configurations. 
We search proxy sets (``w/ cam'' and ``w/o cam'') under different 
availability of the target domain camera annotation. 
} 
\label{table:Result}
\end{table*}

\section{Experiment}
\subsection{Experimental Details}
\textbf{Databases.}~This paper uses a wide range of real-world and synthetic person re-ID datasets. Real-world ones include Market-1501 \cite{zheng2015scalable}, DukeMTMC-reID~\cite{zheng2017unlabeled,ristani2016MTMC}, MSMT17~\cite{wei2018person}, CUHK03~\cite{li2014deepreid}, RAiD~\cite{das2014consistent},
PKU-Reid~\cite{ma2016orientation} and iLIDS~\cite{zheng2009associating}.~Synthetic datasets used are PersonX~\cite{sun2019dissecting}, Randperson~\cite{wang2020surpassing} and UnrealPerson~\cite{zhang2020unrealperson}. Some important details of these datasets are shown in Table~\ref{table:Datasets}. From these datasets, we can select one as the source and another one as the target. The rest will form the data pool (Section~\ref{sec:search a set}).  
When creating the data pool, we only use a portion of identities and their corresponding images. This limits the problem size in our searching process, while preventing dominating the data pool with images in a few datasets. Overall, we consider a total number of 8,144 identities in our data pool.

\textbf{Models to be ranked.} 
We consider $28$ representative baselines and approaches in person re-ID,  including ID-discriminative embedding (IDE)~\cite{zheng2016mars}, part-based convolution baseline (PCB)~\cite{sun2018beyond}, and record 10 different versions of each model during their training procedure. 
For hyper-parameters, we follow their original settings (see supplementary material for more details of the models). 
In total, we have 280 models, \ie, $N=280$ in $\left \{ \mathbf{m}_{i} \right \}_{i=1}^{M}$. All the models are trained from scratch on the source domain. 

\textbf{Searched proxies.}
In this work, we choose the hyper-parameters for proxy set searching as $\lambda=0.6$ for the weighting factor and $K=20$ for the cluster number. The number of identities of the searched proxy sets is set to $500$ (see Section~\ref{sec:parameter}). For more details on the searched proxy sets, please refer to the supplementary materials. We perform search with one RTX-2080TI GPU and a 16-core AMD Threadripper CPU @ 3.5Ghz.

\begin{figure*}[t]
\begin{center}
 	\includegraphics[width=\linewidth]{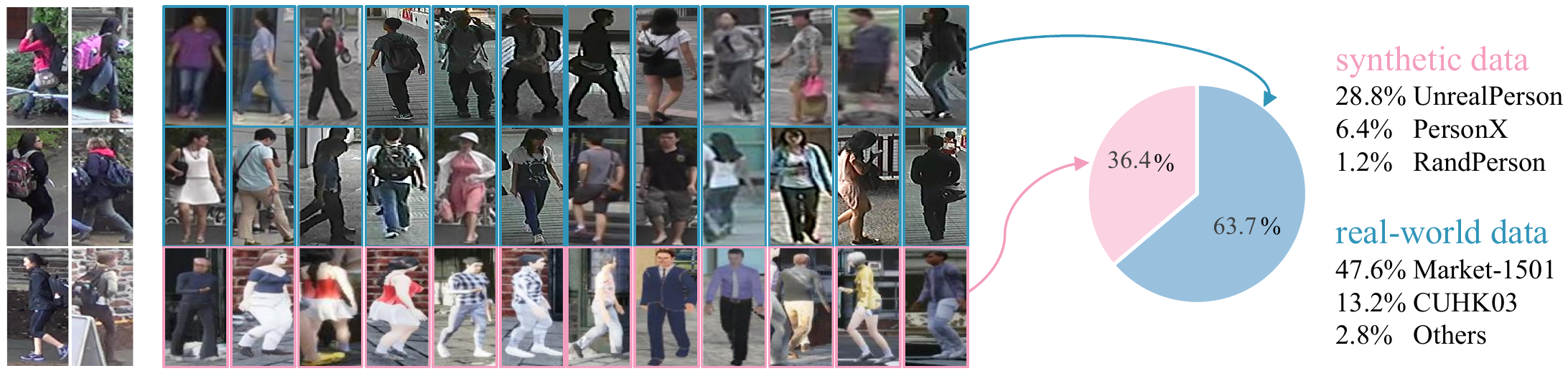}
\end{center}
\caption{Image samples and composition statistics of the searched proxy (MSMT17 as source and Duke as target). \textbf{Left}: unlabeled target; \textbf{Middle}: searched proxy; \textbf{Right}: composition statistics of the searched proxy. 
We observe that the searched proxy overall displays similar lighting and color schemes compared with the target. }
\label{fig:example}
\end{figure*}

\subsection{Evaluation of the Proposed Method}

In Table ~\ref{table:Result}, we compare  the quality of our searched proxy with alternative proxy choices, including individual labeled datasets (datasets in Table~\ref{table:Datasets}), engine-based synthetic images \cite{yao2019simulating}, GAN-based generated images \cite{choi2018stargan,zhong2018generalizing}, pseudo labels on the target validation \cite{fan2018unsupervised}, and a random sample from all the individual labeled dataset (denoted as ``Random'' in Table~\ref{table:Result}). We have the following observations.

\textbf{Effectiveness of the searched proxy over individual datasets.}
Our main observation is that the searched proxy is very competitive to individual datasets as proxy. When MSMT17 is used as source and DukeMTMC-reID is used as target, the searched proxy (``w/o cam'' in Table~\ref{table:Result}) achieves very good ranking correlations ($\rho=0.858$ and $\tau=0.713$), outperforming both individual datasets and other methods by at least $+0.021$ of $\rho$ and $+0.060$ of $\tau$. Similar results can also be found when Market-1501 is selected as target, where the proposed searching method achieves $\rho=0.884$ and $\tau=0.715$, outperforming every alternative by at least  $+0.010$ of $\rho$ and $+0.024$ of $\tau$. 


Besides, the search method is better than a random combination of individual datasets. As shown in Table~\ref{table:Result}, ``Random'' might lag behind some of the better performing individual datasets by up to $-0.343$ of $\rho$ and $-0.300$ of $\tau$. Our searching method, on the other hand, constantly outperforms this random combination by at least $+0.133$ of $\rho$ and $+0.176$ of $\tau$, while achieving competitive or even better results to the individual datasets. 

\begin{figure*}[t]
\begin{center}
 	\includegraphics[width=\linewidth]{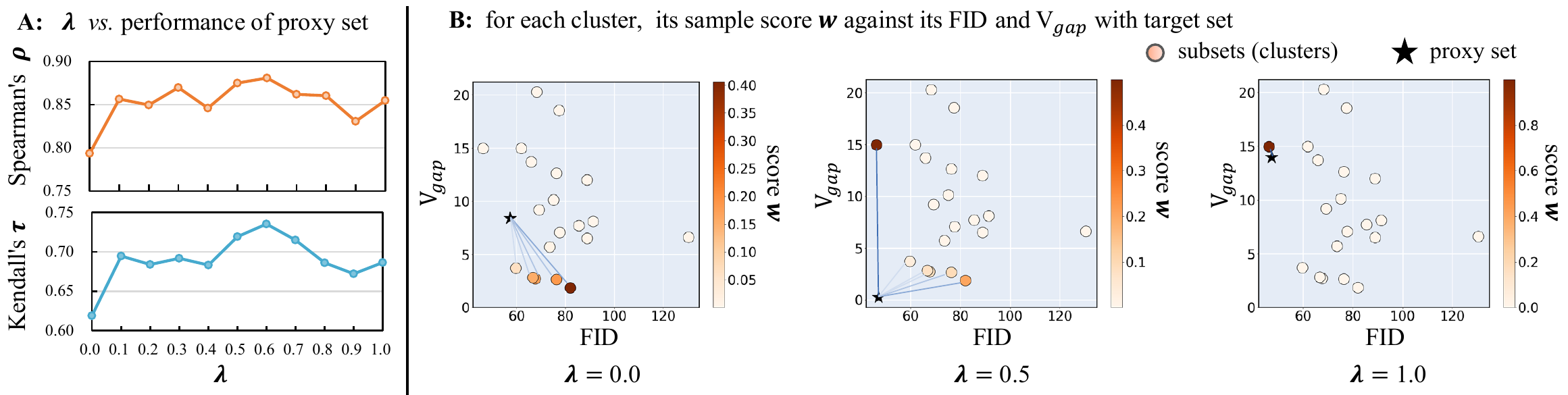}
\end{center}
\caption{The impact of weighting factor $\lambda$ in Eq.~\ref{eq:4}. \textbf{A}: 
searched proxy quality under different $\lambda$ values. 
Then overall sample score $w_{k}$ only considers variance gap when $\lambda=0$, and only considers FID when $\lambda=1$. 
\textbf{B}: sampling scores for each cluster and their contribution to the searched proxy set $\star$ under different $\lambda$ values. Deep colors denote higher sampling scores for the clusters (dots) and higher contributions to the searched proxy (lines).  
Here, the cluster number $K$ is set to $20$. MSMT17 and DukeMTMC-reID are used as source and target, respectively.
}
\label{fig:lambda}
\end{figure*}

\textbf{Utilizing camera annotations of the target domain yields the best performance of proxy.} For example, when MSMT17 and DukeMTMC-reID is used as source and target, repetitively, using camera information in our searching approach further improves the overall proxy quality (ranking correlations) to $\rho=0.882$ and $\tau=0.735$. This shows the advantage of a task-centered searching method design, which aligns well with the cross-camera matching the nature of the person re-ID problem.

\begin{table*}
	\begin{minipage}{0.65\linewidth}
		\centering
		\includegraphics[width=\linewidth]{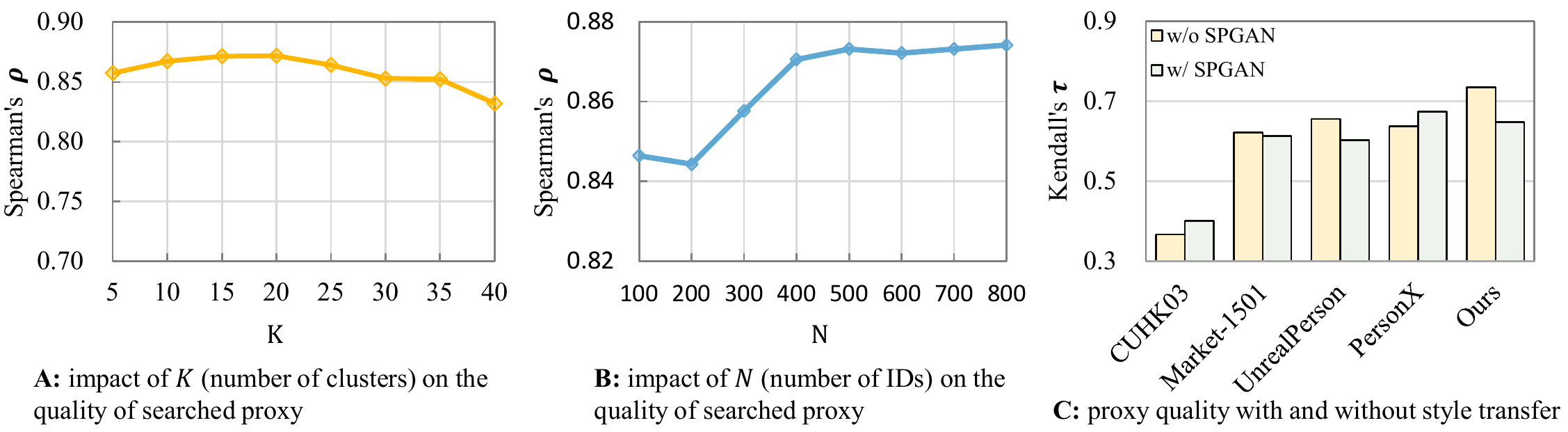}
		\captionof{figure}{The impact of \textbf{A:} number of clutters $K$, \textbf{B:} number of IDs $N$ and \textbf{C:} style transfer on the quality of proxy set. (MSMT17 and DukeMTMC-reID are used as source and target sets, respectively.) }
		\label{fig:spgan}
	\end{minipage}
	\hfill
	\begin{minipage}{0.33\linewidth}\small
		\centering
		    \setlength{\tabcolsep}{0.5mm}{
		\begin{tabular}{l | c| c| c}
			\toprule
		Training	Data  & R1 & R5 & mAP\\
			\midrule
		MSMT17    &   58.0  & 71.6 & 36.5 \\
		Market-1501  &    42.1  & 56.1&  23.9 \\
	    Synthetic data~\cite{yao2019simulating}   &  21.2& 39.7 &  13.5    \\
		Pseudo-label~\cite{fan2018unsupervised} & 67.5 & 80.5&  50.4   \\
		Ours (searched proxy) & 47.9 & 63.2 & 27.9\\
			\bottomrule
		\end{tabular}}
	    \caption{Performance on DukeMTMC-reID using different training sets. Here, R$k$ means rank-$k$ accuracy.
	    }
		\label{table:student}
	\end{minipage}\hfill
\end{table*}

\textbf{Study of the composition of proxy set.}
In Fig.~\ref{fig:example}, we examine the composition of the searched proxy set. With the MSMT17 dataset as source and the DukeMTMC-reID dataset as target, the searching process ends up using more (63.7\%) real-world data compared to synthetic ones, since the real-world data might look more similar to the real-world target of DukeMTMC-reID. Overall, our searched proxy sampled IDs and images look similar to those in the target domain in terms of lighting and colors.

\textbf{Comparison with generated images and pseudo label methods.}
Methods designed for generating or synthesizing training data are found less efficient as proxy sets. For example, engine-based synthesis~\cite{yao2019simulating} achieves a  $\rho$ of 0.756 and $\tau$ of 0.569 for rank correlations ( MSMT17 as source and DukeMTMC-reID as target). This trails behind not only our searched proxy set but also some of the better-performing individual datasets as proxies. 
As for GAN-based method~\cite{zhong2018generalizing} and pseudo label method \cite{fan2018unsupervised}, both of them create image-label pairs using a network, which might introduce inaccurate labels (network decided image label pairs are less reliable than annotated ones). For this reason, the rank correlations of these methods are also sub-optimal.

\textbf{Computational cost in searching a proxy.} 
As shown in Fig. \ref{fig:pip}, there are three steps involved in our proxy searching process. When the MSMT17 is used as source and Market-1501 is used as target, feature extraction and clustering cost about 200 seconds. Then, it takes about 188 seconds to calculate the FIDs and variance gaps. Time for the image sampling process can be neglected. So our algorithm consumes about 400 seconds in total. 
When camera annotations are available, the searching process has no additional cost in the first step. In fact, feature extraction and clustering results can be reused. The overall searching process takes about 1772 seconds for all 6 cameras in the target set.

\subsection{Parameter Analysis}
\label{sec:parameter}

\textbf{Weighting factor $\lambda$ for sampling score.} $\lambda$ encodes the trade-off between FID and $\mathrm{V_{gap}}$ when calculating the sampling score. As shown in Fig. \ref{fig:lambda} \textbf{A}, setting $\lambda$ to 0.6 (as in our current design) gives the best overall quality of the proxy set (highest Spearman's and Kendall's correlation coefficient). Using only either FID or variance gap (setting $\lambda$ to 1 or 0) leads to a quality drop of the searched proxy set. Interestingly, only using FID provides slightly better results compared to only using variance gap. One possible reason is that the FID also considers covariance during computation, which might have a slight overlap with the variance gap. In this case, the variance gap is also reduced when only minimizing FID, which might provide a slight edge to the variant that only uses FID over only using variance gap. 

For more intuitive understandings, we find that only considering variance gap ($\lambda=0$) creates a proxy set that has an even higher variance gap compared to the clusters that majorly contribute to the proxy (Fig.~\ref{fig:lambda} \textbf{B} $\lambda=0.0$). Only considering FID ($\lambda=1$) samples samples mainly from only one cluster, and results in a proxy that is very similar in terms of FID (Fig.~\ref{fig:lambda} \textbf{B} $\lambda=1.0$). When jointly considering both FID and variance gap ($\lambda=0.5$), the resulting proxy has an even lower FID and variance gap compared to the clusters that contribute to it, further indicating the effectiveness of the proposed method (Fig.~\ref{fig:lambda} $\lambda=0.5$ ). 

\textbf{Numbers of clusters $K$ and IDs $N$ of Proxy Set.} 
The proposed method clusters the data pool into $K$ groups based on their ID-averaged features and samples $N$ identities to build the proxy set. Here, we further investigate the influence of the cluster number $K$ and the identity number $N$ on the searched proxy quality. As shown in Fig.~\ref{fig:spgan} \textbf{A}-\textbf{B}, we find that 1) either a too small or too large $K$ can lead to slightly poor proxy quality (here, $N$ is set to 400) and 2) when N gradually becomes larger, the result tends to be stable, so we set the cluster number $K$ to an intermediate value 20, and ID number to 500, to provide relatively good results.

\subsection{Further Understandings}

\textbf{Can we improve the proxies by style transfer?} 
Pixel-level alignment~\cite{deng2018similarity,wei2018person,liu2019adaptive} is commonly used to reduce the domain gap by transferring the image style of one domain into that of the other. For different proxy sets (individual datasets or searched ones), we employ SPGAN~\cite{deng2018similarity} to translate them into the style of the target domain. 
We present the correlation coefficients in Fig.~\ref{fig:spgan} \textbf{C}.
Taking the DukeMTMC-reID dataset as target data, we transfer several proxy sets to DukeMTMC-reID style through SPGAN~\cite{deng2018similarity} and use the style-transferred proxy sets to ranking models. 
It is found that SPGAN cannot bring consistent improvements to the model ranking proxies. 
Despite these mixed results, we note that the best performance is still held by the searched proxy (without style transfer). 

\textbf{Can we train re-ID models on proxy sets for a certain target?} In Table~\ref{table:student}, we find that directly applying re-ID models (IDE~\cite{zheng2016mars}) trained on the searched proxy set does not lead to competitive performance on the target domain, despite the fact that the proxy set is searched for that target specifically. In comparison, pseudo-label~\cite{fan2018unsupervised}, a method that underperforms our method in building proxy sets for model ranking, actually achieves the best result in building training sets for domain adaptation models. This suggests that our problem is quite different from training data search, although they might appear similar at first glance.

\textbf{Effectiveness of MMD in replacing FID.} 
We replace FID with MMD in Eq.~\ref{eq:4}, which is another way to calculate the distribution difference between two datasets. We use MSMT17 and DukeMTMC-reID as source and target, respectively. 
We observe that replacing FID with MMD yields $-0.0056$ Spearman’s $\rho$ and $-0.0161$ Kendall’s $\tau$, suggesting that MMD has a similar effect with FID.

\section{Conclusion}
This paper studies an important and practical problem: when some source models are directly applied to an unseen target domain, can we rank their performance without having to know the ground-truth labels for (a representative subset of) the target domain?  We answer this question by using a so-called {\em target proxy} for un-referenced model evaluation. We first propose a number of baseline approaches, \ie, using the source data as proxy, or using various cross-domain datasets as proxy. We analyze the underlying reasons for the (in)effectiveness of such baselines and identify that the domain gap and diversity gap are two important factors affecting the quality of a proxy. We therefore adopt a search strategy that uses a weighted combination of these two metrics as objective. Experiments on public person re-ID datasets validate our strategy and let us gain rich insights into dataset similarity and model generalization.

\section*{Acknowledgement}
This work was supported in part by the ARC Discovery Early Career Researcher Award (DE200101283), the ARC Discovery Project (DP210102801). Hongdong Li's research is also funded in part by an Australian Research Council  Discovery  grant (DP 190102261).

{\small
\bibliographystyle{ieee_fullname}
\bibliography{egbib}
}

\clearpage
\section*{Appendix}
In the supplementary material, we 1) include the details of models ranked in the main paper, 2) provide experimental results when the DukeMTMC-reID dataset is used as the source, 3) provide more visual examples of the searched proxy sets, and 4) provide further discussion.

\section{Person Re-identification Models}
The main paper uses 280 models for ranking, which come from $28$ representative baselines and approaches in person re-ID. These methods/models are selected from three popular Github repositories: Person\_reID\_baseline\footnote{\url{https://github.com/layumi/Person_reID_baseline_pytorch}}, reid-strong-baseline\footnote{\url{https://github.com/michuanhaohao/reid-strong-baseline}} and deep-person-reid\footnote{\url{https://github.com/KaiyangZhou/deep-person-reid}}. Furthermore, for each method, we record 10 different versions corresponding to different epochs during the training process. Therefore, a total of $28\times10=280$ models are used. 

\begin{figure}
\begin{center}
	\includegraphics[width=\linewidth]{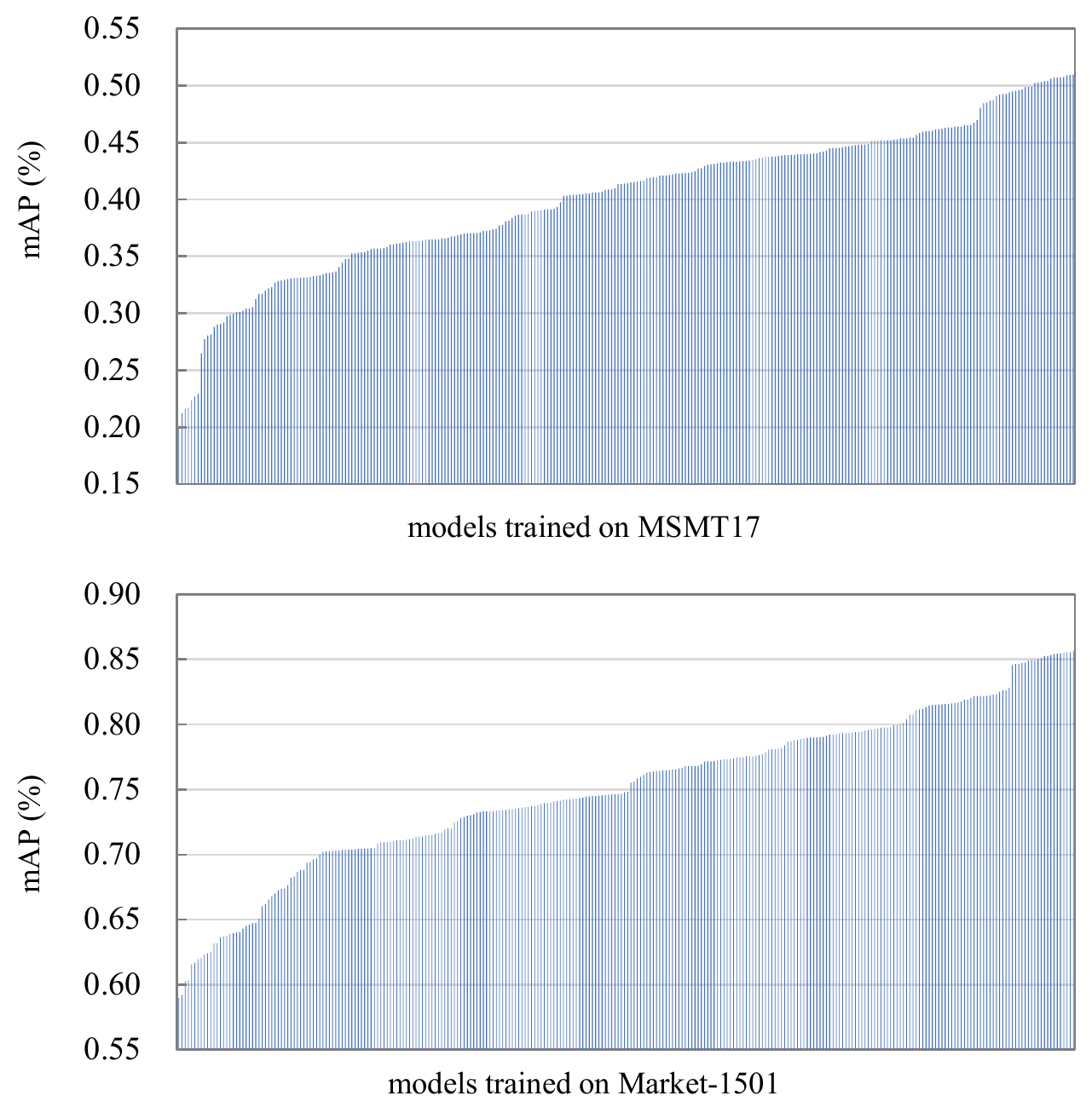}
\end{center}
\caption{mAP ($\%$) scores of 280 models trained and tested on the same dataset: \textbf{A:} MSMT17 and \textbf{B:} Market-1501.
}
\label{fig:models}
\end{figure}

The names of the 28 methods are shown in Table~\ref{table:models}. Note that, although some methods use the same CNN architecture, such as ResNet50, their model accuracies are different because they use different training strategies or hyper-parameters (\emph{e.g.,} learning rate and the dimension of the FC layer output). Fig.~\ref{fig:models} shows the mAP scores of the 280 models when trained and tested on a given dataset, such as the MSMT17 or Market-1501. Results show that these models have different image representation ability for person re-ID, so ranking them is feasible to reflect their relative representing performance on both target and proxy set.

Although the mAP scores of some models may be the same on a certain dataset, it will not influence the rank correlation evaluation since Kendall's tau can draw accurate generalizations for rankings with repeated rank~\cite{akoglu2018user}.

\section{DukeMTMC-reID as Source}
Table~\ref{table:Result-sup} compares the quality of proxy sets in terms of Spearman's $\rho$ and Kendall’s $\tau$ when the DukeMTMC-reID and Market-1501 datasets are used as source and target, respectively. The result have similar trends to those in the main paper. For example, a weak correlation between the source and target sets is shown by the rank correlation coefficients $\rho=0.314$ and $\tau=0.225$. Further, the UnrealPerson dataset, when used as proxy, has higher correlation values of $\rho=0.837$ and $\tau=0.668$ with the target than the other individual datasets. Comparing with individual proxy sets and proxies generated by other methods, our proxy sets have higher rank correlation coefficients with the target set.

\begin{table}[t]\footnotesize
\begin{center}
    \setlength{\tabcolsep}{1mm}{
        \begin{tabular}{l|l|l} 
            \Xhline{1.2pt}
            \multicolumn{1}{l|}{Person\_reID\_baseline}    &    \multicolumn{1}{l|}{reid-strong-baseline} &     \multicolumn{1}{l}{deep-person-reid}  \\
            \hline
\tabincell{l}{IDE, PCB, \\DenseNet \\IDE-lr0.05, \\PCB-lr0.02,\\ DenseNet-lr0.05, \\ IDE-fix-bn, \\PCB-fix-bn, \\ DenseNet-fix-bn}& \tabincell{l}{ResNet18, ResNet34 \\
ResNet50, ResNet101, \\ ResNet152, SeResNet50, \\SeResNet101, SeResNet152, \\ SeResNeXt50, \\softmax, softmax-triplet, \\ softmax-triplet-with-center, \\ IBN-Net50-a}
& 
\tabincell{l}{osnet-x0-25, \\osnet-x0-50, \\ osnet-x0-75, \\osnet-x1-0, \\osnet-x1-0-cosinelr, \\resnet50-fc512, \\ resnet50}
\\
            \Xhline{1.2pt}             
    \end{tabular}}
\end{center}
    \caption{Names of methods that are used for model ranking in the main paper. ``lr'' represents learning rate.} \label{table:models}
\end{table}

\begin{table*}[t]\small
\begin{center}
    \setlength{\tabcolsep}{0.5mm}{
        \begin{tabular}{l| l | l |c c c c c c c |c c c c|c c} 
            \Xhline{1.2pt}
   \multicolumn{1}{c|}{\multirow{2}{*}{Source}} &   \multicolumn{2}{c|}{\multirow{2}{*}{Target}}  &\multicolumn{7}{c|}{Individual Dataset} & \multicolumn{4}{c|}{Other Dataset Generation Methods} & \multicolumn{2}{c}{Ours}\\ 
 \cline{4-16}
 \multicolumn{1}{c|}{} & \multicolumn{2}{c|}{}  & \scriptsize{CUHK03} & \scriptsize{Duke} & \scriptsize{Market} & \scriptsize{MSMT17} & \scriptsize{RandPerson} & \scriptsize{PersonX} & \scriptsize{UnrealPerson} & \scriptsize{Random} & \tiny{Attr. descent~\cite{yao2019simulating}} & \tiny{StarGAN~\cite{choi2018stargan}} & \tiny{pseudo-label~\cite{fan2018unsupervised}} & w/o cam & w/ cam \\ 
 \Xhline{1.2pt}
  \multirow{2}{*}{Duke} & \multirow{2}{*}{Market}& 
       $\rho$ & 0.568 &0.314 & -&0.835 & 0.745 &0.705 & 0.837&0.642 & 0.574 & 0.741& 0.827 & \textbf{0.866}& \blue{\textbf{0.893}}\\
  &  & $\tau$ & 0.400 &0.225 & -&0.646 & 0.568 &0.519 & 0.668 &0.504 & 0.424 & 0.562& 0.623 & \textbf{0.698}&\blue{\textbf{0.706}}\\
\Xhline{1.2pt} 
    \end{tabular}}
\end{center}
\caption{Comparison of different proxy sets when using DukeMTMC-reID as source and  Market-1501 as target.} 
\label{table:Result-sup}
\end{table*}

\begin{figure*}
\begin{center}
	\includegraphics[width=\linewidth]{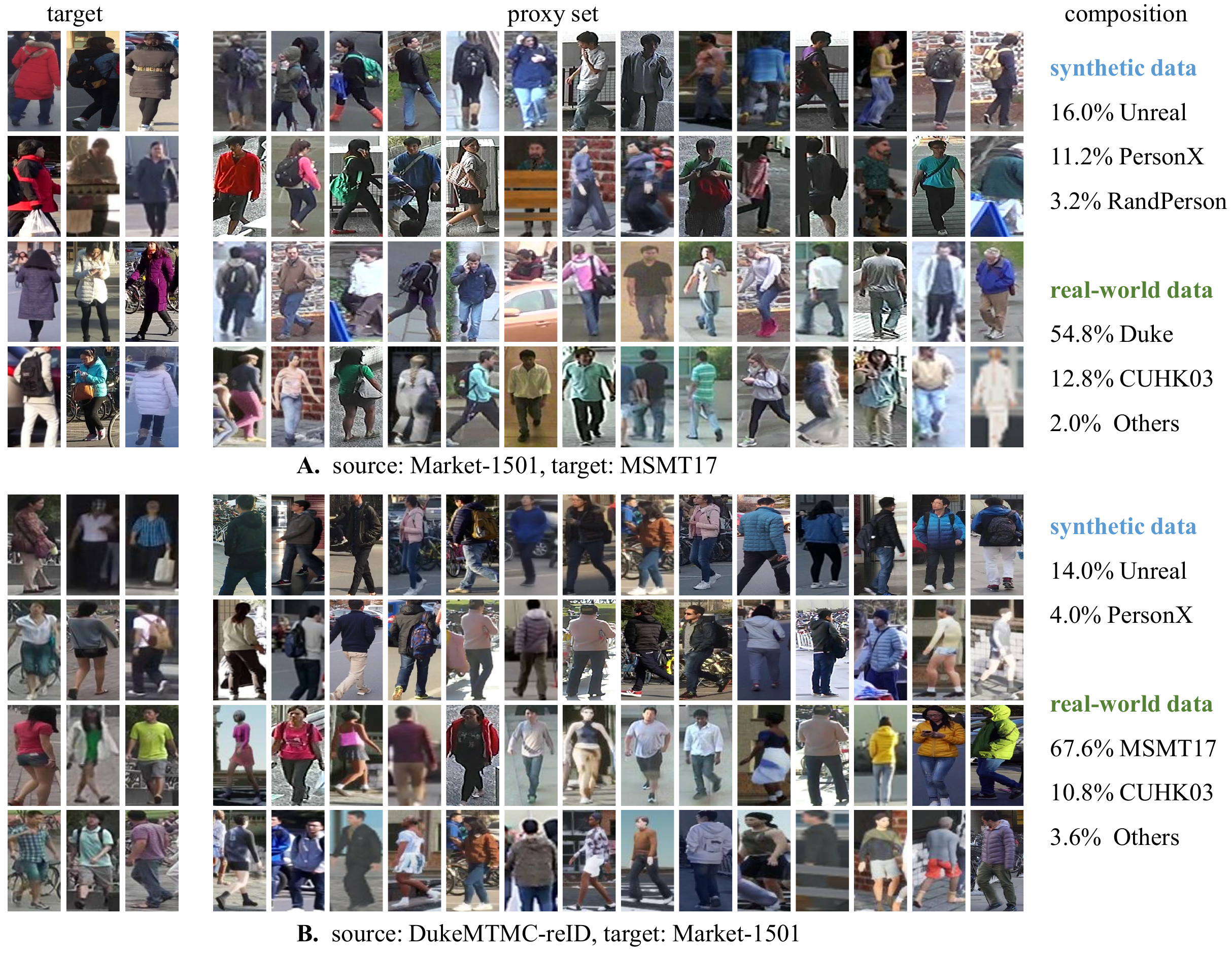}
\end{center}
\caption{Image samples and compositions of searched proxy sets for different source and targets sets. 
}
\label{fig:images}
\end{figure*}

\section{Image Samples of Proxy Sets}
Fig.~\ref{fig:images} shows the image samples and composition statistics of the searched proxy sets. We observe that the proposed method finds images with similar styles with the target, such as background color and illumination. For example, the searched images have various illumination conditions, and the illumination in MSMT17 also exhibits such characteristics (Fig.~\ref{fig:images} \textbf{A}). Further, we observe that real-world data take up a larger proportion (\eg, about 70$\%$ when using MSMT17 as the target) than synthetic data in the composition of the searched set. A possible reason is that real-world images have a small domain gap with the real-world target data.

\section{Further Discussion}

\begin{figure}
	\begin{center}
 			\includegraphics[width=1\linewidth]{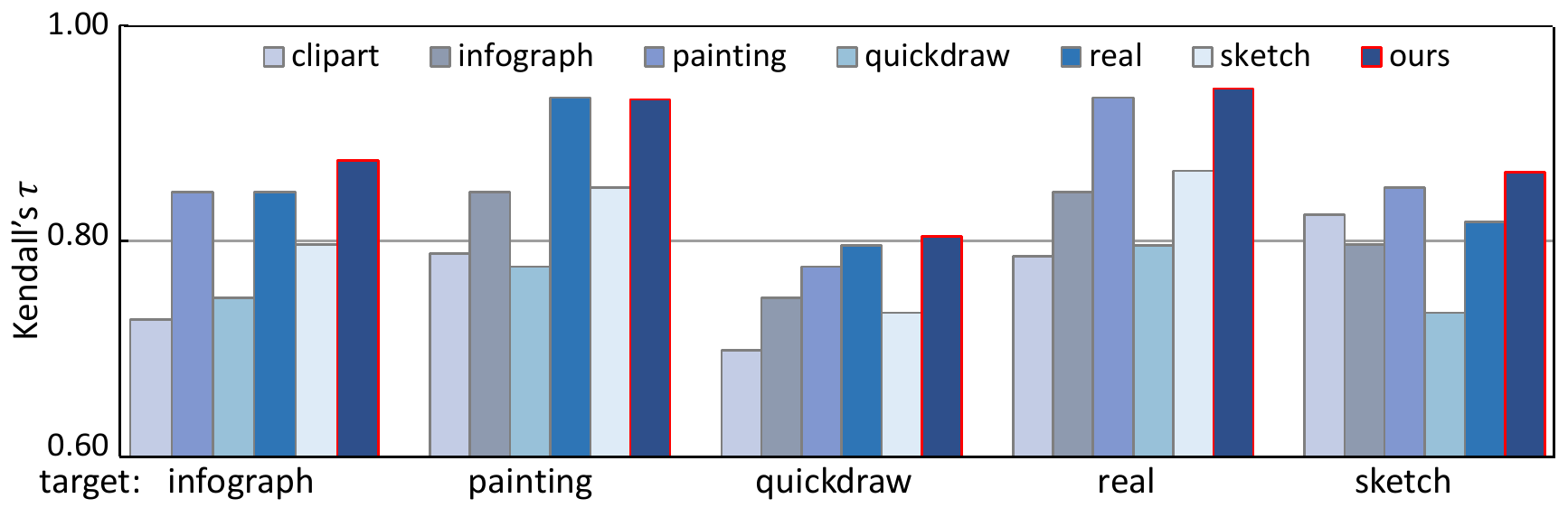}
	\end{center}
	\caption{Comparison of different proxy sets on different targets of the DomainNet datasets over 70 models. Kendall’s Rank Correlation $\tau$ is used as metric.}
\label{fig:classification}
\end{figure}

\textbf{Can the proposed method generalize to other tasks?} 
We discuss this question on image classification task by using the DomainNet dataset, which has 6 domains, \textit{i.e.,} Clipart, Infograph, Painting, Quickdraw, Real and Sketch, and 345 categories. We took Clipart as source and the other 5 in turn as target. The results are shown in Fig.~\ref{fig:classification}. Our searched proxy achieves best results on four out of five targets and second best on the other target (painting). 

The results suggest that 1) the chosen searching metric (FID and variance gap) is also effective in classification, 2) the proposed method a potential solution for ranking models of other tasks, such as image classification. However, most other tasks expect re-ID would require additional assumptions for evaluation, making the choices in candidate datasets limited based on existing datasets. For example, image classification requires the source and target domains to have the same classes. Therefore, the largest domain adaptation dataset, DomainNet, might still be sub-optimal for investigating this problem because 1) it only offers 4 datasets (besides source and target) to construct the database pool, and 2) the distributions of the 4 domains (\emph{e.g.,} sketch, real) are tremendously different.

Above limitations prevent our method from giving a clear margin over individual datasets are proxy, because the target will be approximated by mainly sampling images from one candidate rather than multiple. We will include above discussion and further study this problem by collecting data of other tasks in our future work. 

\textbf{Best models selected by proxy sets.}
Table~\ref{table:map of models} shows the mAP scores of the best models selected by different proxy sets (MSMT17 as source and DukeMTMC-reID as target).

\begin{table}[h]\small
	\begin{center}
		   \setlength{\tabcolsep}{1mm}{
		\begin{tabular}{l c c c c c c}
			\hline
		proxy	  &  \scriptsize{Market}&  \scriptsize{UnrealPerson} & \scriptsize{pseudo-label} &\scriptsize{random} &\scriptsize{ours} & \scriptsize{oracle} \\
		\hline
     mAP (\%) & 36.98 & 37.16 & 36.70 &36.05 & 38.10& 38.12 \\
			\hline
		\end{tabular}}
	\end{center}
	\caption{mAP scores of best models selected by proxy sets. }
\label{table:map of models}
\end{table}

The model selected by the searched proxy (our) has the best performance on the target set, verifying the effectiveness of our approach.

\end{document}